\documentclass[conference]{IEEEtran}
\IEEEoverridecommandlockouts

\usepackage{cite}
\usepackage{amsmath,amssymb,amsfonts}
\usepackage{algorithmic}
\usepackage{graphicx}
\usepackage{subfig}
\usepackage{textcomp}
\usepackage{xcolor}
\usepackage{multirow}
\usepackage{amsmath}
\usepackage{amsfonts,amssymb}
\usepackage{booktabs}
\usepackage{stfloats}
\usepackage{bbding}
\usepackage[letterpaper, top=0.75in, bottom=1.05in, left=0.625in, right=0.625in, columnsep=0.25in]{geometry}

\def\BibTeX{{\rm B\kern-.05em{\sc i\kern-.025em b}\kern-.08em
    T\kern-.1667em\lower.7ex\hbox{E}\kern-.125emX}}
\begin{document}

\title{Communication-Efficient Federated Domain Generalization in Edge Networks via Dynamic Prototype Regulation}

\author{
Yuankun Xia$^{\dagger}$, 
Hui Wang$^{\dagger}$$^{*}$, Ling Yuan$^{\dagger}$, Ziyu Hong$^{\dagger}$, Di Yin$^{\dagger}$  and Yufeng Zhou$^{\dagger}$ \\
\textit{$^{\dagger}$School of Computer Science and Technology, Zhejiang Normal University, Jinhua 321004, China} \\ 
\textit{\{yuankunx, dzxxyl, hongziyu, yandi1, nebulaweek\}@zjnu.edu.cn, hwang@zjnu.cn} \\
$^{*}$\textit{Corresponding author: Hui Wang}
}

\maketitle

\begin{abstract}
Federated learning (FL) is a distributed paradigm for Internet of Things (IoT) edge networks. However, deploying Vision-Language Models (VLMs) over wireless edge networks presents distinct challenges: large model parameters increase uplink transmission latency over bandwidth-constrained fading channels, and non-independent and identically distributed (Non-IID) data exacerbates client drift. Existing feature alignment solutions often introduce auxiliary computations that consume time slots allocated for wireless parameter transmission, increasing the likelihood of straggler dropouts under system delay constraints. To address these issues, we propose DPGR-GCAM, a communication-efficient federated domain generalization framework. By employing a dual-parameter-efficient fine-tuning (PEFT) architecture, we design a channel-aware tiered transmission mechanism that adaptively truncates payloads to maximize node survival rates under fading conditions. To mitigate client drift, we introduce the Gradient-Conflict Aware Modulation (GCAM) mechanism. GCAM operates within the low-dimensional feature manifold to dynamically resolve multi-objective optimization conflicts with minimal edge computation latency, thereby reducing computation-induced network outages. Experiments demonstrate that DPGR-GCAM improves cross-domain generalization and sustains convergence under network fading scenarios with a 10\% client participation rate.
\end{abstract}


\section{Introduction}
The deployment of 6G and the Internet of Things (IoT) has led to an increase in multimodal data generated by mobile edge devices. To utilize this data while maintaining privacy and reducing backhaul transmission, Federated Learning (FL) serves as a distributed collaborative training paradigm\cite{FL}. Recently, Vision-Language Models (VLMs) have shown capability in zero-shot generalization\cite{VLM}. However, deploying these foundation models over wireless edge networks requires transmitting large parameter sets, which increases uplink transmission latency over bandwidth-constrained fading wireless channels. Under system maximum tolerable delay ($T_{max}$) constraints, these payloads can lead to communication outages, reducing the active client participation rate and affecting global convergence.

In realistic wide-area IoT scenarios, heterogeneous edge nodes exhibit non-independent and identically distributed (Non-IID) data characteristics\cite{noniid1,noniid2,noniid3}. Domain shifts caused by varying geographical locations and environments contribute to client drift. In conventional federated domain generalization (FedDG) frameworks\cite{feddg}, solutions that address feature space fragmentation often require additional local computation overhead. From a network perspective, increased local computation latency ($T_{k}^{comp}$) reduces the available time for wireless parameter transmission, which increases the vulnerability of edge devices to channel fading and straggler issues.

Parameter-efficient fine-tuning (PEFT)\cite{peft} methods exchange lightweight parameters between clients and the server, reducing the communication burden. However, some PEFT-based federated mechanisms (e.g., PromptFL\cite{promptfl} and FedCLIP\cite{Fedclip}) lack feature correction tailored for data heterogeneity. Consequently, local models can overfit to private distributions, which limits the generalization capability of the aggregated global model on unseen target domains\cite{feftllm}.

To address the trade-off between communication latency, computation overhead, and cross-domain generalization, we propose DPGR-GCAM. We formulate the deployment of federated VLMs as a joint optimization problem bounded by Shannon capacity, transmit power, and system maximum delay constraints. This approach provides a distributed paradigm designed for fading channels and Non-IID scenarios. The main contributions of this paper are summarized as follows:
\begin{itemize}
    \item We introduce a dual-PEFT architecture in OFDMA federated networks. Combining visual adapters with Gradient-Conflict Aware Modulation (GCAM) mitigates local overfitting and gradient conflicts under Non-IID conditions. This maintains node participation rates under $T_{max}$ limits and reduces cumulative communication overhead.
    \item To address client drift, we design a communication-aware global prototype aggregation scheme. We also propose the GCAM mechanism. Operating within the low-dimensional feature manifold via gradient approximation, it modulates optimization conflicts with minimal computation latency.
    \item Experiments on DomainNet and Office-Caltech10 benchmarks evaluate the cross-domain generalization performance of DPGR-GCAM compared to baseline methods.
\end{itemize}

The remainder of the paper is organized as follows. The System Model and Methodology is presented in Section \ref{sec2}. Then the experiments and analysis in Section \ref{sec3}. Finally, Section \ref{sec4} concludes the paper.

\begin{figure*}[t] 
    \centering
    \includegraphics[width=0.95\textwidth]{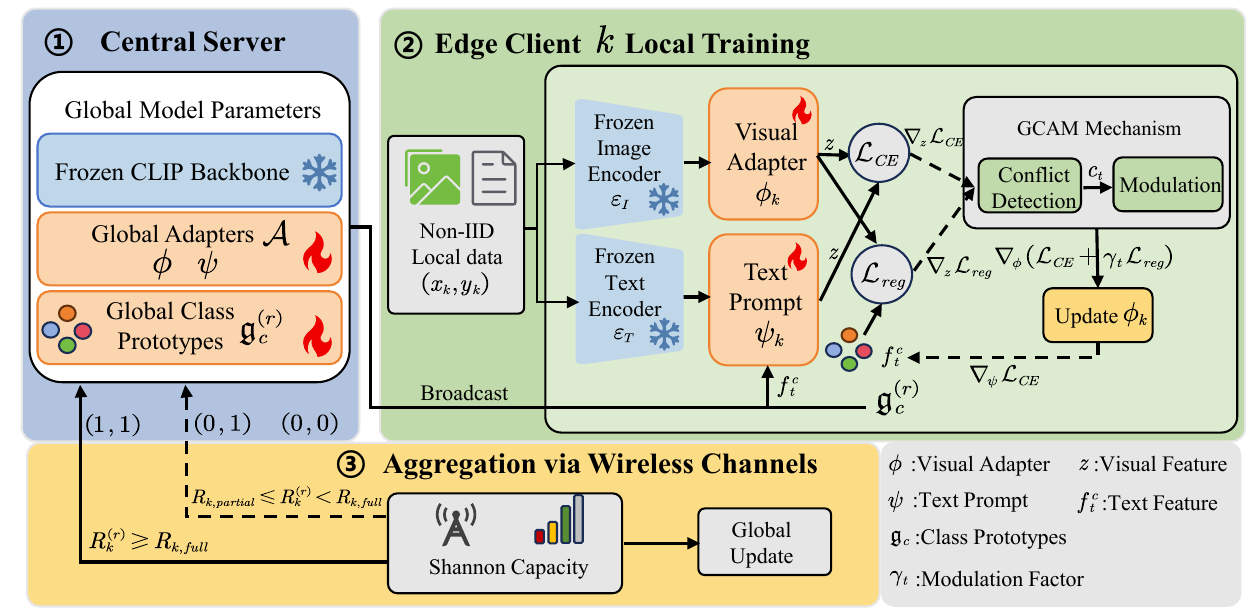} 
    \caption{The overall workflow of the proposed DPGR-GCAM framework for federated domain generalization over wireless edge networks. The system comprises three stages: (1) Global broadcast of lightweight tunable parameters and class prototypes; (2) Local training, where the GCAM mechanism dynamically resolves feature-level optimization conflicts to regulate the visual adapter update; and (3) Channel-aware aggregation, which adaptively truncates the uplink payload based on real-time Shannon capacity to maximize node survival rates under fading channels.}
    \label{fig:overall_workflow}
\end{figure*}
    
\section{System Model and Methodology}
    \label{sec2}
    In this section, we first establish the system model for FedDG over edge networks and introduce the physical-layer wireless communication and computation delay constraints. Subsequently, we present the PEFT architecture based on VLMs. To overcome model drift under extreme data heterogeneity and unreliable wireless links, we propose a communication-aware distributed prototype alignment mechanism, synergized with an innovative GCAM mechanism. Finally, we formulate the entire framework as a joint optimization problem strictly bounded by communication and computation constraints. The overall workflow of the proposed DPGR-GCAM framework is illustrated in Fig.\ref{fig:overall_workflow}.
    \subsection{System Model and Problem Formulation}
    Consider an OFDMA-based federated edge computing network comprising a Central Server and $K$ edge devices, denoted by the set $\mathcal{K}=\{1,2,...,K\}$. Each edge node $k\in\mathcal{K}$ maintains a private local dataset $\mathcal{D}_{k}=\{(x_{i},y_{i})\}_{i=1}^{|\mathcal{D}_{k}|}$, where $x\in\mathcal{X}$ is the input image and $y\in\mathcal{Y}=\{1,...,C\}$ represents the label space. In complex wide-area IoT scenarios, the data generation distributions across different nodes exhibit extreme Non-IID characteristics.

    Wireless Communication and Latency Model: In the $r$-th global communication round, the uplink data transmission rate from edge node $k$ to the central server is given by the Shannon capacity formula:
    \begin{equation}
        R_{k}^{(r)} = B_{k} \log_{2}\left(1 + \frac{P_{k} |h_{k}^{(r)}|^{2}}{N_{0}}\right),
    \end{equation}
    where $B_{k}$ is the physical resource block bandwidth allocated to node $k$, $P_{k}$ is its transmit power, $h_{k}^{(r)}$ denotes the channel fading coefficient in the current round, and $N_{0}$ is the power spectral density of the additive white Gaussian noise (AWGN).

    In edge networks, a single federated aggregation round is strictly constrained by a maximum tolerable delay $T_{max}$. Traditional federated learning typically employs a hard dropout mechanism \cite{drop,drop2}. Under fading channels, this strategy easily triggers knowledge starvation within the global model. To break through this physical bottleneck, capitalizing on the asymmetric parameter characteristics of the Dual-PEFT architecture, we design a Channel-Aware Tiered Transmission mechanism. We decouple the uplink payload into lightweight text prompts $\psi$ alongside relatively larger visual adapters $\phi$ and prototypes $z$. Consequently, we define the minimum rate thresholds for full and partial transmissions as follows:
    \begin{equation}
        R_{k, full} = \frac{|\phi| + |\psi| + |z|}{T_{max} - T_{k}^{comp}}, \quad R_{k, partial} = \frac{|\psi|}{T_{max} - T_{k}^{comp}}.
    \end{equation}

    To maximize the survival rate of nodes under fading channels and formally define the transmission state, we formulate the decoupled communication outage indicator pair $(\mathbb{I}_{k,\phi}^{(r)}, \mathbb{I}_{k,\psi}^{(r)})$ based on the real-time Shannon capacity $R_k^{(r)}$:
    \begin{equation}
        (\mathbb{I}_{k,\phi}^{(r)}, \mathbb{I}_{k,\psi}^{(r)}) = 
        \begin{cases} 
        (1, 1), & \text{if } R_k^{(r)} \ge R_{k, full} \\ 
        (0, 1), & \text{if } R_{k, partial} \le R_k^{(r)} < R_{k, full} \\ 
        (0, 0), & \text{if } R_k^{(r)} < R_{k, partial} 
        \end{cases},
    \end{equation}
    where $\mathbb{I} = 1$ denotes a successful parameter transmission within the time slot, and $\mathbb{I} = 0$ represents a communication outage triggered by the delay constraint. This adaptive payload truncation strategy ensures that edge nodes suffering from severe fading can still successfully upload their highly informative text semantic features ($\psi$). While strictly adhering to physical delay constraints, this approach maximizes the effective participation rate and the system's physical robustness under adverse channel conditions.
    
    \subsection{Communication-Efficient PEFT Architecture}
    Deploying vision foundation models CLIP with robust generalization capabilities on resource-constrained edge devices by directly transmitting and fine-tuning the entire massive parameter set is practically infeasible over realistic wireless communication networks. To this end, we introduce a PEFT architecture to adapt pre-trained knowledge to downstream heterogeneous nodes with extremely low communication overhead.

    In our distributed fine-tuning protocol, the backbone network parameters of the CLIP image encoder $\mathcal{E}_{I}$ and text encoder $\mathcal{E}_{T}$ are strictly frozen across all clients. We introduce only two types of lightweight trainable modules: Adapter parameters $\phi$ inserted in the later stages of the image encoder, and Prompt Learner parameters $\psi$ on the text side. The visual and textual feature representations are extracted as:
    \begin{equation}
        f_v = \mathcal{E}_I(x; \phi), \quad f_t^y = \mathcal{E}_T(y; \psi).
    \end{equation}
    The prediction probability is calculated through the cosine similarity between features:
    \begin{equation}
        p(y=c|x) = \frac{\exp(\cos(f_v, f_{t}^c)/\tau)}{\sum_{j=1}^{C} \exp(\cos(f_v, f_{t}^j)/\tau)}.
    \end{equation}
    Under this architecture, edge nodes are only required to exchange the lightweight parameters $\phi$ and $\psi$. Since $|\phi|+|\psi|\ll |\mathcal{E}_{I}|+|\mathcal{E}_{T}|$, the uplink communication payload $|\Theta_{k}|$ is drastically compressed from hundreds of MB to the KB scale. This exponential payload reduction fundamentally boosts the survival rate of nodes under severe channel conditions, providing a highly viable edge computing foundation for handling domain generalization with high-dimensional image data.

    \subsection{Communication-Aware Distributed Prototype Alignment}
    To fundamentally resolve feature space fragmentation induced by Non-IID data, we propose constructing cross-domain shared global prototypes within the feature manifold. In the $r$-th global communication round, edge node $k$ computes the local prototype $p_{k,c}^{(r)}$ for category $c$ based on its current local fine-tuned parameters:
    \begin{equation}
        p_{k,c}^{(r)}=\frac{1}{|\mathcal{D}_{k,c}|}\sum_{(x,y)\in\mathcal{D}_{k,c}}\mathcal{E}_{I}(x;\phi_{k}^{(r)}).
    \end{equation}
    
    Unlike conventional full-aggregation under ideal network assumptions, the central server maintains two distinct sets of active nodes based on the tiered transmission indicators: the prompt-active set $\mathcal{S}_{r, partial} = \{k \in \mathcal{K} \mid \mathbb{I}_{k,\psi}^{(r)}=1\}$ and the fully-active set $\mathcal{S}_{r, full} = \{k \in \mathcal{K} \mid \mathbb{I}_{k,\phi}^{(r)}=1\}$. Because computing global prototypes strictly requires the uploaded local visual features, the data-volume-weighted prototype aggregation is executed exclusively over the fully-active set $\mathcal{S}_{r, full}$:
    \begin{equation}
        \mathfrak{g}_{c}^{(r)}=\frac{\sum_{k\in\mathcal{S}_{r, full}}|\mathcal{D}_{k,c}|\cdot p_{k,c}^{(r)}}{\sum_{k\in\mathcal{S}_{r, full}}|\mathcal{D}_{k,c}|}.
    \end{equation}
    For edge nodes that only upload partial text semantics due to deep fading, they do not contribute to the current round's visual prototype aggregation but still actively update the global text prompts, thereby preserving partial cross-domain semantic collaboration.
    
    Upon receiving the broadcasted global prototypes, edge nodes introduce explicit spatial geometric constraints. For a sample $x$, feature-level alignment regularization is implemented by minimizing the Euclidean distance between its extracted feature and the corresponding global prototype $\mathfrak{g}_{y}^{(r)}$:
    \begin{equation}
        \mathcal{L}_{reg}(x,y;\mathfrak{g}_{y}^{(r)})=||\mathcal{E}_{I}(x;\phi_{k})-\mathfrak{g}_{y}^{(r)}||_{2}^{2}.
    \end{equation}


    \subsection{GCAM}
    In FedDG, prototype regularization with static weights frequently leads to sub-optimal convergence under severe Non-IID data. We strictly formulate the local update as a bi-objective optimization problem, balancing the local cross-entropy loss $\mathcal{L}_{CE}$ and the prototype alignment loss $\mathcal{L}_{reg}$.

    In wireless edge networks, conventional second-order gradient computations drastically increase the local computational burden, causing a surge in computation latency $T_{k}^{comp}$ and highly elevating the risk of triggering $T_{max}$ timeouts. To achieve conflict perception with minimal computational overhead, we propose Feature-level Gradient Approximation. Let $z=\mathcal{E}_{I}(x;\phi)$ denote the low-dimensional image feature embedding. We mathematically bridge the parameter space and the low-dimensional feature manifold $z=\mathcal{E}_{I}(x;\phi)$ through the following lemma.

    Drawing upon gradient routing \cite{fisher,gradient}, we approximate high-dimensional parameter-space optimization conflicts via a well-conditioned Jacobian mapping to formulate computationally lightweight feature-level gradients:
    \begin{equation}
        g_{CE}^{(t)}=\nabla_{z}\mathcal{L}_{CE}, \quad g_{reg}^{(t)}=\nabla_{z}\mathcal{L}_{reg}.
    \end{equation}
    We employ cosine similarity as the conflict metric factor $c_{t}\in[-1,1]$:
    \begin{equation}
        c_{t}=\frac{\langle g_{CE}^{(t)},g_{reg}^{(t)}\rangle}{||g_{CE}^{(t)}||_{2}\cdot||g_{reg}^{(t)}||_{2}}.
    \end{equation}
    Since the feature dimension is astronomically smaller than the adapter parameter size ($d \ll |\phi|$), this approximation gracefully reduces the computational complexity of conflict detection from $\mathcal{O}(|\phi|)$ to $\mathcal{O}(d)$, enabling dynamic penalty modulation with zero marginal edge computation overhead. Based on $c_{t}$, GCAM designs an elegantly simple, parameter-free dynamic modulation formula:
    \begin{equation}
        \gamma_{t}=\gamma_{0}\cdot\frac{r}{R}\cdot(1-c_{t}),
    \end{equation}
    where $\gamma_{0}$ is the base strength constant and $R$ is the total number of communication rounds. When the model drifts towards a local optimum and induces a gradient conflict $(c_{t}<0)$, the system adaptively imposes stronger global constraints, effectively eliminating the reliance on heuristic hyperparameter tuning.

    \subsection{Joint Optimization Problem Formulation}
    Synthesizing the lightweight PEFT architecture, the network-state-aware prototype aggregation, and the computationally minimalist GCAM modulation, the proposed joint optimization problem for federated domain generalization over edge networks is strictly formulated as a multi-constraint optimization problem. Bounded by channel bandwidth, device transmit power, and system maximum latency, the rigorous mathematical expression is explicitly detailed as:
    \begin{equation}
        \begin{aligned}
                \min_{\{\phi, \psi, \mathbb{I}, P, B\}} \;& \frac{1}{\sum_{r=1}^{R} |\mathcal{S}_{r}|} \sum_{r=1}^{R} \sum_{k \in \mathcal{S}_{r}} \big[ \mathcal{L}_{CE} + \gamma_{t}\mathcal{L}_{reg} \big]\\
                \text{s.t.} \;\;
                & \text{C1: } T_{k}^{comp} + \frac{1}{R_{k}^{(r)}} \Big( |\phi_{k}|\mathbb{I}_{k,\phi}^{(r)} + |\psi_{k}|\mathbb{I}_{k,\psi}^{(r)} \\
                & \qquad\qquad\qquad\qquad + |z_{k}|\mathbb{I}_{k,\phi}^{(r)} \Big) \le T_{max} \\
                & \text{C2: } {\textstyle \sum_{k \in \mathcal{K}}} B_{k} \le B_{total}, \;\; \text{C3: } 0 < P_{k} \le P_{max} \\
                & \text{C4: } \mathbb{I}_{k,\phi}^{(r)}, \mathbb{I}_{k,\psi}^{(r)} \in \{0, 1\}, \quad \forall k \in \mathcal{K}, \forall r
        \end{aligned}
    \end{equation}

    Problem (12) is a Mixed-Integer Non-Linear Programming (MINLP) problem due to the binary transmission indicators (C4) and the non-convex cross-entropy loss. Simultaneously solving this entails unacceptable computational overhead for IoT devices. Therefore, we tackle this intractable physical boundary through a system-level decoupled strategy: the physical-layer communication constraints (C1-C4) are heuristically satisfied via the Channel-Aware Tiered Transmission mechanism, while the multi-objective learning conflict is dynamically resolved by GCAM. This guarantees strict adherence to the $T_{max}$ limits without requiring heavy resource allocation solvers, achieving a dynamic equilibrium between local empirical risk and global semantic alignment under adverse network conditions.

    \section{Experiments And Analysis}
    \label{sec3}
    \subsection{Experimental Setting}
    To comprehensively evaluate the effectiveness of the proposed DPGR-GCAM framework over bandwidth-constrained and fading IoT edge networks, we conduct extensive system-level FedDG simulations under realistic physical-layer constraints.
    
    \textbf{Wireless Network and System Settings:} We consider an OFDMA-based federated edge network scenario. In this scenario, $K$ edge devices are uniformly distributed within a cell radius of 500 meters from the central server. The wireless channel is modeled using Rayleigh fading, where the channel gain $|h_k^{(r)}|^2$ follows an exponential distribution with unit mean. The path loss is modeled as {\small$ PL(d) = 128.1 + 37.6 \log_{10}(d)$} in dB, where $d$ is the distance in kilometers. The total system bandwidth is 20 MHz, dynamically and equally allocated among all active devices. The maximum transmit power of edge nodes $P_{max}$ is set to 20 dBm, and the AWGN power spectral density $N_0$ is set to -114 dBm/MHz. Crucially, to emulate strict edge computing and transmission bottlenecks, the maximum tolerable system delay per federated aggregation round $T_{max}$ is set to 5.0 seconds. Any device failing to complete local computation and parameter upload within $T_{max}$ is strictly adjudged as a straggler and directly dropped according to our physical-layer outage indicator function $\mathbb{I}_{k}^{(r)}$.
    
    \textbf{Datasets and Data Heterogeneity:} We employ the Office-Caltech10\cite{OC10} benchmark and the large-scale multi-domain dataset DomainNet\cite{DN} for cross-domain research. To characterize and simulate the heterogeneous data distribution of edge nodes in federated networks, we utilize a Dirichlet distribution $\text{Dir}(\alpha)$ governed by a concentration parameter $\alpha$. We set $\alpha \in \{0.01, 0.5\}$, where $\alpha=0.01$ simulates extreme label skew, and $\alpha=0.5$ represents a moderate heterogeneity level commonly encountered in realistic federated collaborations.
    
    \textbf{Baselines and Implementation Details:} We benchmark DPGR-GCAM against a suite of state-of-the-art federated learning methods. The baselines encompass traditional federated alignment algorithms (FedAvg, FedProto), as well as the latest PEFT federated collaboration strategies designed for large VLMs, including PromptFL, FedCLIP, FedAPT, and FedOTP. Across all VLM-based baselines and our framework, we uniformly adopt the pre-trained CLIP ViT-B/16 as the backbone network, strictly freezing all underlying parameters of both the image and text encoders. During local fine-tuning, clients solely update the lightweight visual adapters and text prompt learners. The local training epochs are set to 5, the global communication rounds $R$ to 50, and optimization is driven by SGD with a base learning rate of 0.01 and a batch size of 32. For the GCAM mechanism, the base regularization strength $\gamma_0$ is set to 1.0.
    
    \begin{table*}[t] 
      \centering
      \vspace{0.05in}
      \caption{Cross-Domain Generalization Accuracy (\%) on DomainNet and Office-Caltech10 under Varying Data Heterogeneity ($\alpha$)}
      \label{tab:performance_evaluation}
      \begin{tabular}{@{}ll ccccccc ccccc@{}} 
        \toprule
        \multirow{2}{*}{\textbf{$\alpha$}} & \multirow{2}{*}{\textbf{Methods}} & \multicolumn{7}{c}{\textbf{DomainNet}} & \multicolumn{5}{c}{\textbf{Office-Caltech10}} \\
        \cmidrule(lr){3-9} \cmidrule(lr){10-14} 
        & & c & i & p & q & r & s & \textbf{avg} & a & w & d & c & \textbf{avg} \\
        \midrule
        \multirow{7}{*}{0.01} 
              & FedAvg\cite{fedavg}   & 81.37 & 59.28 & 77.32 & 26.69 & 89.29 & 75.53 & 68.25 & 93.75 & 89.66 & 93.55 & 93.78 & 92.69 \\
              & FedProto\cite{fedproto} & 83.68 & 65.54 & 76.79 & 26.86 & 88.38 & 76.50 & 69.63 & 94.79 & 93.10 & 93.55 & 95.33 & 94.19 \\
              & FedCLIP\cite{Fedclip}  & 82.99 & 58.60 & 76.53 & 25.83 & 89.26 & 75.16 & 68.06 & 95.83 & 93.10 & 93.55 & 93.33 & 93.95 \\
              & PromptFL\cite{promptfl} & \textbf{85.94} & 38.36 & 62.79 & 21.71 & 79.96 & 67.39 & 59.36 & 96.35 & 97.56 & 98.98 & 94.22 & 96.78 \\
              & FedAPT\cite{FedAPT}   & 82.69 & 39.69 & 59.31 & 21.60 & 77.34 & 65.59 & 57.70 & 97.91 & 98.28 & \textbf{100.00} & \textbf{96.44} & \textbf{98.16} \\
              & FedOTP\cite{fedotp}   & 77.09 & 63.17 & 48.60 & 28.11 & 61.80 & 74.63 & 58.90 & 95.31 & 98.28 & \textbf{100.00} & 94.67 & 97.07 \\
              & \textbf{Ours} & 85.48 & \textbf{65.39} & \textbf{79.70} & \textbf{38.94} & \textbf{90.27} & \textbf{81.55} & \textbf{73.56} & \textbf{95.83} & \textbf{100.00} & 96.77 & \textbf{96.44} & 97.26 \\
        \midrule
        \multirow{7}{*}{0.5} 
              & FedAvg\cite{fedavg}   & 82.06 & 59.23 & 77.27 & 28.20 & 89.50 & 76.39 & 68.78 & 94.79 & 92.83 & 95.77 & 94.67 & 94.52 \\
              & FedProto\cite{fedproto} & 84.25 & 67.24 & 80.16 & 28.63 & 90.57 & 78.08 & 71.49 & 95.31 & 93.41 & 93.67 & 96.10 & 94.82 \\
              & FedCLIP\cite{Fedclip}  & 82.66 & 58.60 & 76.68 & 25.43 & 89.38 & 75.37 & 68.02 & 95.92 & 94.23 & 94.12 & 95.33 & 94.90 \\
              & PromptFL\cite{promptfl} & 87.86 & 53.00 & 64.90 & 22.54 & 83.04 & 69.78 & 63.52 & 97.39 & 98.27 & \textbf{100.00} & 95.11 & 97.69 \\
              & FedAPT\cite{FedAPT}   & 83.85 & 49.01 & 61.72 & 18.20 & 80.25 & 66.41 & 59.91 & 97.32 & 97.85 & \textbf{100.00} & 95.33 & 97.63 \\
              & FedOTP\cite{fedotp}   & 77.59 & 63.87 & 49.34 & 28.71 & 62.10 & 75.33 & 59.49 & 95.31 & \textbf{100.00} & \textbf{100.00} & 92.89 & 97.05 \\
              & \textbf{Ours} & \textbf{88.13} & \textbf{69.45} & \textbf{82.69} & \textbf{40.54} & \textbf{92.28} & \textbf{83.14} & \textbf{76.04} & \textbf{97.92} & \textbf{100.00} & \textbf{100.00} & \textbf{97.78} & \textbf{98.93} \\
        \bottomrule
      \end{tabular}
    \end{table*}
    
    \subsection{Main Results}
    \textbf{Cross-Domain Generalization Analysis.} Table \ref{tab:performance_evaluation} reports the average generalization accuracy under varying degrees of data heterogeneity ($\alpha \in \{0.01, 0.5\}$). DPGR-GCAM establishes significant performance superiority on the large-scale DomainNet dataset. For instance, under moderate heterogeneity ($\alpha=0.5$), our framework achieves an average accuracy of 76.04\%, outperforming conventional regularization baselines. Critically, while recent VLM federated methods reduce communication payloads, they suffer from severe performance degradation under extreme domain shifts, sometimes even underperforming standard FedAvg. This corroborates that blindly grafting PEFT modules without communication-aware feature correction induces severe local overfitting. Conversely, by resolving optimization conflicts within the low-dimensional manifold at near-zero marginal latency, GCAM enables DPGR-GCAM to achieve an optimal dynamic equilibrium between local fitting and global semantic alignment.
    \begin{figure}[t] 
      \centering
      \subfloat[Office-Caltech10]{
        \includegraphics[width=0.47\columnwidth]{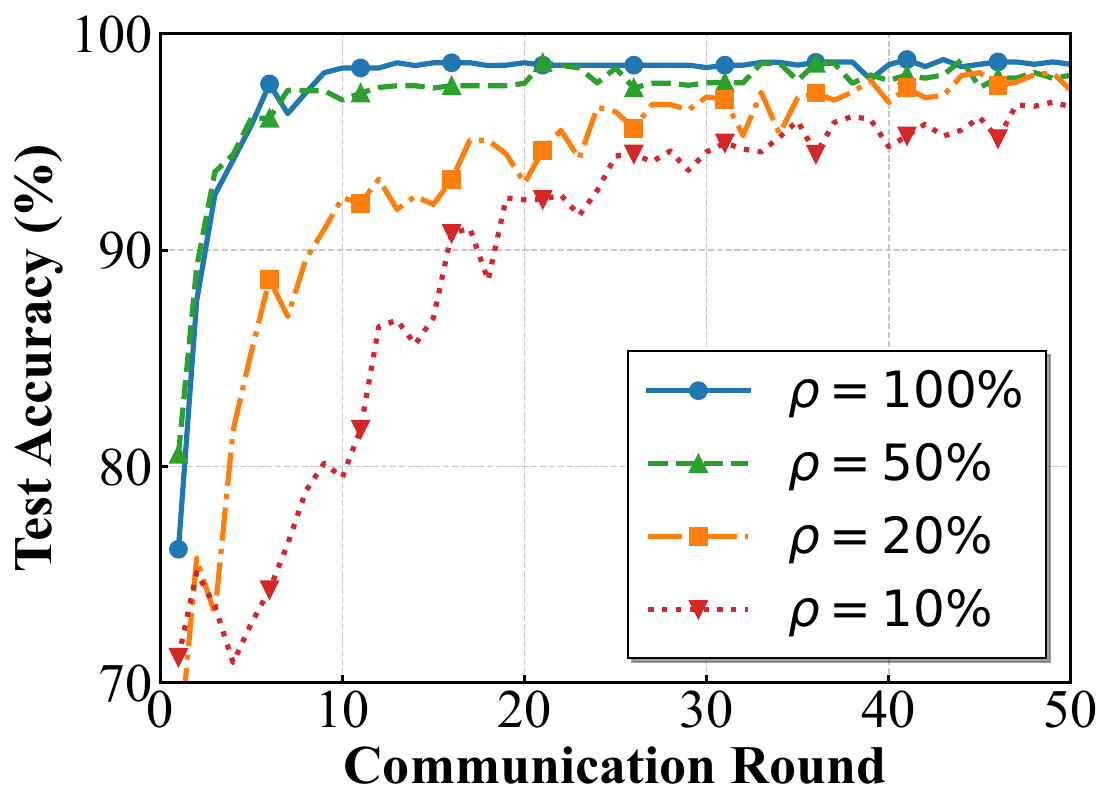} 
        \label{fig:sub_a}
      }
      \hfil 
      \subfloat[DomainNet]{
        \includegraphics[width=0.47\columnwidth]{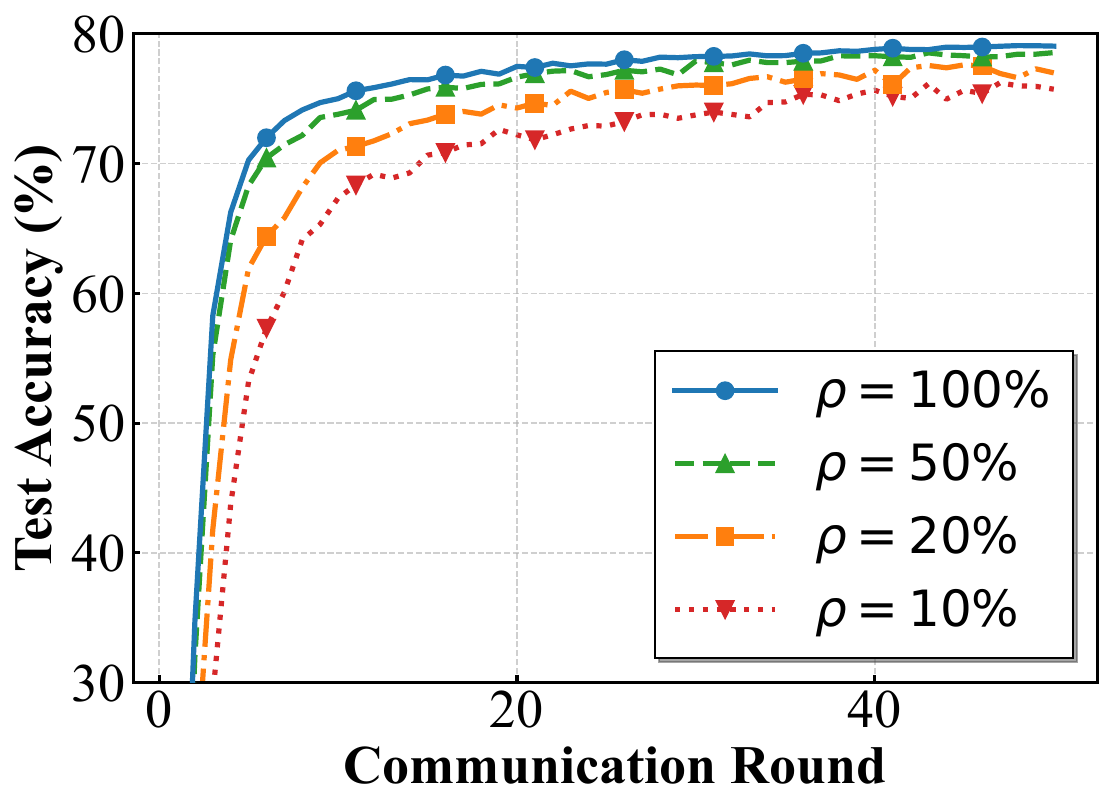} 
        \label{fig:sub_b}
      }
      \caption{Robustness evaluation under intermittent connectivity on different datasets. (a) Moderate-scale Office-Caltech10. (b) Large-scale DomainNet.}
      \label{fig:robustness_eval} 
    \end{figure}    
    
    \textbf{System Resource Overhead Analysis.} In severely resource-constrained IoT networks, massive communication payloads and memory footprints constitute the core bottlenecks precipitating system delay violations. Table II profiles the physical resource overhead of different frameworks. The conventional Full Fine-tuning protocol mandates an uplink transmission payload of 487.04 MB and occupies nearly 11 GB of peak memory on edge nodes. Over Rayleigh fading channels, this inevitably exceeds $T_{max}$, driving the survival rate to $\rho\approx0\%$ and triggering collaboration collapse. By adopting our designed dual-PEFT architecture, DPGR-GCAM slashes the uplink communication payload by 97.4\%, fundamentally alleviating bandwidth congestion over the physical wireless channels. Compared to baselines like FedOTP that rely on computationally intensive Optimal Transport operations, DPGR-GCAM strictly confines dynamic prototype alignment within the low-dimensional feature space. The computation and memory overhead introduced by this architectural design is negligible, enabling safe and highly efficient deployment on standard 8GB IoT edge nodes.
    
    \begin{table}[t] 
      \centering
      \caption{Communication and Computation Overhead on Edge Nodes}
      \label{tab:overhead_analysis}
      \begin{tabular}{@{}l c c c@{}}
        \toprule
        \multirow{2}{*}{\textbf{Method}} & \textbf{Trainable} & \textbf{Comm. Payload} & \textbf{Peak Memory} \\
        & \textbf{Params (M)} & \textbf{(MB)} & \textbf{(MB)} \\
        \midrule
        Full Fine-tuning & 127.68 & 487.04 & 10,968 \\
        PromptFL\cite{promptfl}         & 2.83   & 10.78  & 6,124  \\
        FedAPT\cite{FedAPT}           & 3.09   & 11.78  & 6,127  \\
        FedOTP\cite{fedotp}           & 5.65   & 10.78  & 6,147  \\
        \textbf{Ours} & \textbf{3.35} & \textbf{12.78} & \textbf{6,125} \\
        \bottomrule
      \end{tabular}
    \end{table}

    \begin{table}[t]
      \centering
      \caption{Effectiveness of Key Components in DPGR-GCAM}
      \label{tab:ablation_study}
      \begin{tabular}{@{}l c c@{}}
        \toprule
        \textbf{Components} & \textbf{DomainNet} & \textbf{Office-Caltech10} \\
        \midrule
        PEFT only                  & 71.45 & 95.23 \\
        PEFT + $\mathcal{L}_{reg}$ & 72.84 & 96.43 \\
        w/o Adapter                & 74.16 & 97.21 \\
        \textbf{DPGR-GCAM (Full)}  & \textbf{76.04} & \textbf{98.93} \\
        \bottomrule
      \end{tabular}
    \end{table}
    
    \textbf{System Robustness Analysis.}  In practical 6G/IoT deployments, edge nodes frequently fail to complete tasks within the strict $T_{max}$ time slot and degrade into stragglers due to deep wireless channel fading or computational bottlenecks. Fig.\ref{fig:robustness_eval} illustrates the convergence performance of the system under varying effective client participation rates ($\rho \in \{10\%, 20\%, 50\%, 100\%\}$), strictly governed by the outage indicator function $\mathbb{I}_{k}^{(r)}$. Experimental results indicate that the DPGR-GCAM framework exhibits system-level fault tolerance against extreme physical network degradation. On the massive DomainNet dataset, the framework secures a highly competitive accuracy of 75.73\% even at $\rho=10\%$. This corroborates that our dynamic prototype-guided mechanism effectively suppresses client drift triggered by communication outages, cementing its status as a highly robust solution for volatile edge computing environments.
    
    
    \textbf{Ablation Studies.} Table \ref{tab:ablation_study} provides system ablation studies to systematically disentangle the performance contributions of our core architectural designs. Relying exclusively on parameter-efficient fine-tuning (PEFT only) yields a sub-optimal accuracy of 95.23\% on Office-Caltech10, as the local model is highly susceptible to drift induced by Non-IID data. Upon integrating static prototype-guided regularization (PEFT + $\mathcal{L}_{reg}$), the performance distinctively leaps to 96.43\%, verifying the efficacy of anchoring local optimization trajectories with global semantics. Furthermore, if the visual adapter is removed (w/o Adapter), the system degenerates into pure text prompt tuning, causing the accuracy to regress to 97.21\%. This directly validates the systemic necessity of the dual-PEFT architecture in facilitating multi-modal domain adaptation. Ultimately, the full DPGR-GCAM achieves a peak accuracy of 98.93\% on Office-Caltech10 and secures an absolute advantage of 76.04\% on the massive DomainNet dataset, substantiating the superiority of this joint architectural design.

    \section{Conclusion}
    \label{sec4}
    In this paper, we proposed DPGR-GCAM, a communication-efficient and hardware-friendly FedDG framework tailored for wireless edge networks. To overcome the severe transmission latency penalties incurred by deploying large VLMs over bandwidth-constrained and fading channels, we introduced a dual-PEFT architecture. By strictly freezing the backbone network, this architecture reduces the uplink communication payload by 97.4\%. Furthermore, to combat client drift induced by extreme data heterogeneity, we designed the GCAM mechanism. GCAM dynamically resolves multi-objective optimization conflicts within a low-dimensional feature manifold at minimal marginal edge computation latency, effectively mitigating network outages triggered by computation overload. Extensive system-level simulations demonstrate that DPGR-GCAM achieves competitive generalization robustness and convergence speed, even under extreme network outage scenarios where only 10\% of clients complete aggregation. This work establishes a highly resilient distributed paradigm for deploying multimodal foundation models under stringent 6G and IoT physical constraints.
    
    \section*{Acknowledgment}
    
    This research was supported by the National Natural Science Foundation of China under Grant No. 62171413 and the Zhejiang Central Science and Technology Innovation Corridor Joint Fund of Zhejiang Provincial Natural Science Foundation of China under Grant No. LJHSD26F020001.
    
    \bibliographystyle{IEEEtran}
    \bibliography{refs}
    
    \end{document}